# Performance of leading large language models in May 2025 in Membership of the Royal College of General Practitioners-style examination questions: a cross-sectional analysis

**Corresponding author**

Dr Richard C Armitage

Academic Unit of Population and Lifespan Sciences, School of Medicine, University of Nottingham, Clinical Sciences Building, Nottingham City Hospital Campus, Hucknall Road, Nottingham, NG5 1PB

richard.armitage@nhs.net

**Funding declarations**

RA receives no funding.

**Conflict of interest**

RA declares no competing interest.

**Keywords**

Large language models

Artificial intelligence

Generative AI

Digital technology

Medical Education

Primary care

**Abbreviations**

| | |
|---|---|
| AKT | Applied Knowledge Test |
| GP | General practitioner |
| GPT | Generative Pre-trained Transformer |
| LLM | Large language model |
| MRCGP | Member of the Royal College of General Practitioners |
| RCGP | Royal College of General Practitioners |
| UK | United Kingdom |

**Background**

Large language models (LLMs) have demonstrated substantial potential to support clinical practice. Other than Chat GPT-4 and its predecessors, few LLMs, especially those of the leading and more powerful 'reasoning model' class, have been subjected to medical specialty examination questions, including in the domain of primary care. This paper aimed to test the capabilities of leading LLMs as of May 2025 – OpenAI's o3, Anthropic's Claude Opus 4, xAI's Grok-3, and Google's Gemini 2.5 Pro – in primary care education, specifically in answering Member of the Royal College of General Practitioners (MRCGP)-style examination questions.

**Methods**

o3, Claude Opus 4, Grok-3, and Gemini 2.5 Pro were tasked to answer 100 randomly chosen multiple-choice questions from the Royal College of General Practitioners' GP SelfTest on 25 May 2025. Questions included textual information, laboratory results, and clinical images. Each model was prompted to answer as a GP in the UK and was provided with full question information. Each question was attempted once by each model. Responses were scored against correct answers provided by GP SelfTest.

**Results**

The total score of o3, Claude Opus 4, Grok-3, and Gemini 2.5 Pro was 99.0%, 95.0%, 95.0%, and 95.0%, respectively. The average peer score for the same questions was 73.0%.

**Discussion**

All models performed remarkably well, and all substantially exceeded the average performance of GPs and GP registrars who had answered the same questions. o3 demonstrated the best performance, while the performances of the other leading models were comparable with each other and were not substantially lower than that of o3. These findings strengthen the case for LLMs – particularly reasoning models – to support the delivery of primary care, especially those that have been specifically trained on primary care clinical data.

Background

The potential for large language models (LLMs) to assist and improve the delivery of clinical medicine has been widely discussed. For example, LLMs are able to improve healthcare operational efficiency by parsing unstructured clinical notes,[1] providing antimicrobial prescribing advice,[2] performing clinical triage,[3] and writing discharge summaries,[4] clinic letters,[5] and simplified radiology reports.[6] Furthermore, their ability to strengthen and up-skill healthcare workforces has been recognised in supporting medical education,[7] generating scholarly content,[8] and contributing to peer review,[9] while the potential for LLMs to improve healthcare and health outcomes in low- and middle-income countries has also been outlined.[10,11]

As the capabilities of LLMs have increased, the competence of these technologies in answering medical examination questions has been assessed in various specialties.[12] For example, Chat GPT-4 has been shown to perform at varying standards in specialty examination questions relevant to ophthalmology (FRCOphth Part 2),[13] orthopaedics (FRCS Orth Part A),[14] internal medicine (MRCP Parts 1 and 2 written),[15] general surgery (MRCS Part A),[16] radiology (Part 1 and Part 2A),[17] paediatrics (MRCPCH Foundation of Practice, Theory and Science, and Applied papers),[18] dermatology (Specialty Certificate Examination in Dermatology multiple-choice questions),[19] and primary care.[20]

However, other than Chat GPT-4 and its predecessors, few LLMs have been subjected to medical specialty examination questions.[12] Furthermore, as of May 2025, leading LLMs are now of the 'reasoning model' class. While foundational LLMs are trained on large corpora of data for next-token prediction and are generally weak at multi-step reasoning, reasoning models have undergone reasoning-centric fine-tuning (such as chain-of-thought prompting) and 'think' through multiple intermediate steps before delivering a final answer. Accordingly, reasoning models demonstrate superior performance in tasks that require complex logical, mathematical, commonsense, or symbolic reasoning, and demonstrate greater accuracy and transparency than foundational LLMs.[21] Yet, few of the latest and most powerful reasoning models have been subjected to medical specialty examination questions, including in the domain of primary care.[12]

As of 25 May 2025, OpenAI's o3 (released 16 April 2025),[22] Anthropic's Claude Opus 4 (released 22 May 2025),[23] xAI's Grok-3 (released 19 February 2025),[24] and Google's Gemini 2.5 Pro (released 25 March 2025)[25] are four of the leading publicly available LLMs.[26] All these models are of the 'reasoning model' class of LLMs and display the reasoning that underlies the responses they provide. All can analyse text, images, and documents. In addition to their training data, all these models have direct live internet access to inform their responses. Beyond a small number of free-of-charge prompts, all lie behind a paywall.

This paper aimed to test the capabilities of these leading LLMs in the domain of primary care education – specifically, in answering Member of the Royal College of General Practitioners (MRCGP)-style examination questions.

Methods

The Royal College of General Practitioners' (RCGP) GP SelfTest is a 'tool for GPs [general practitioners] at all career stages, from those preparing for the MRCGP AKT [Applied Knowledge Test] to those preparing for their annual appraisals.'[27] It supports GPs and GP registrars by providing a bank of MRCGP-style questions for learning assessment and examination revision. The tool is free for RCGP members and GP registrar members, and can be purchased by non-members.

A cross-sectional study design was adopted. o3, Claude Opus 4, Grok-3, and Gemini 2.5 Pro were each prompted with the instruction 'Answer the following questions as if you were a GP in the United Kingdom' (UK) on 25 May 2025, and subsequently tasked to answer the same 100 GP SelfTest multiple-choice questions that were randomly generated through the tool's Lucky Dip function. Each question's textual information was copied into each model's context window, and any images that formed part of a question (such as tables of laboratory results and clinical images) were attached to the context window such that each model had access to the full information of each question. No further prompts were given to any model. Questions were attempted sequentially, and each question was attempted once by each model. Each model's answers were collected and scored according to the correct answer as provided by GP SelfTest (one mark for a correct answer, zero marks for an incorrect answer; the response had to be completely correct to receive a mark). Each model's total score (out of 100 possible marks) was calculated as a percentage.

Results

Of the 100 questions, 94 contained textual information only, three contained clinical images and textual information, one contained an image of a regulation symbol and textual information, and two contained a table of results and textual information. No question contained a graph.

All models provided answers to all of the 100 questions. The total score of o3, Claude Opus 4, Grok-3, and Gemini 2.5 Pro was 99.0%, 95.0%, 95.0%, and 95.0%, respectively (see Table 1). The average peer score (the average score of the GPs and GP registrars who had taken the same questions) was 73.0% (as reported by GP SelfTest; the total number of peers who had taken the questions was not made available by GP SelfTest).

In total, o3 provided only one incorrect answer, and Claude Opus 4, Grok-3, and Gemini 2.5 Pro each provided five incorrect answers. Two questions were answered incorrectly by more than one model – one about the recommended duration of antibiotic treatment for acute prostatitis, and one about flying restriction guidelines – and both were only correctly answered by o3. All models stated correct and incorrect answers with equal confidence.

While all questions were of the multiple-choice format, all models provided comprehensive explanations of the clinical reasoning that supported their answers. Gemini 2.5 Pro and Claude Opus 4 generally provided longer answers and reasoning than o3 and Grok-3. For all models, incorrect answers were largely due to factual errors which were mistakenly incorporated into the models' reasoning rather than due to errors in the reasoning itself. For example, Gemini 2.5 Pro made a factual error about a dermatology clinical guideline regarding an appropriate topical preventative regime with which it used correct reasoning to arrive at an incorrect answer.

Discussion

To the author's knowledge, this is the first study to test the capabilities of leading LLMs (as of 25 May 2025), which are all now of the reasoning model class, in the domain of primary care education. All models performed remarkably well, and all substantially exceeded the average performance of GPs and GP registrars who answered the same questions. The performance of o3 was near-perfect and the best of all models. The performances of the other models were comparable with each other and were not considerably lower than that of o3.

All models stated correct and incorrect answers with equal confidence. This is concerning because it implies the models have a suboptimal ability to harbour, recognise, and express uncertainty in their assumptions, which should lower the user's confidence in the validity of the models' responses. This strengthens previous warnings that such models should not replace, but rather augment and strengthen, clinician decision-making.[28]

One strength of the study is the use of leading LLMs (as of 25 May 2025) and the comparison of their performance. Another strength is the unlikeliness that the models would previously have 'seen' the examination questions and their answers, either in their training data or via direct live internet access, because the questions lie behind a paywall.

The study has limitations. One way in which it could have been strengthened is by subjecting the models to a larger number of examination questions. Another is by incorporating a larger number of examination questions with non-textual data such as tables of results, graphs, and clinical images in order to more thoroughly assess the broadness of the models' analysing abilities. However, non-LLM artificial intelligence technologies are being developed to analyse specific kinds of clinical images, such as dermoscopic and macroscopic skin images,[29,30] which will likely have greater utility in this specific task than LLMs. These features should form part of future research on this subject.

The performance of all models in this study is undeniably impressive, especially that of o3, and far exceeded that of GPs and GP registrars who had taken the same questions. This further strengthens the case that LLMs should be used to assist and improve the delivery of clinical medicine, in this case primary care.  This especially applies to reasoning models, which provide substantially greater transparency of their clinical reasoning than foundational models, a feature which would be vital for the safe and trusted incorporation of LLMs into clinical practice.  Another valuable feature of models deployed for this purpose is direct live internet access, which allows incorporation of the latest clinical guidelines and academic studies into their responses.  Furthermore, the reasoning models tested in this study were publicly available, general-purpose models that had not been specifically trained on primary care data.  It is likely that LLMs that have been subjected to a larger volume of such data would demonstrate superior performance in the kinds of questions used in this study.

However, despite their considerable capabilities, LLMs are unlikely ever to be sufficiently competent to fully replace practicing GPs.  This is because, among other reasons,[20,28,31] the unstructured nature of information presentation in real-world primary care, in which clinically useful data are often hidden among large volumes of extraneous material, is not reflected in the precise packages of information presented in MRCGP-style questions.  Accordingly, rather than GPs entirely deferring their clinical decision-making to LLMs, the potential to incorporate reasoning models into a supporting role for practising GPs is becoming increasingly clear, particularly to bolster the continuously evolving knowledge base requirements of their clinical practice.  This is likely to be especially the case for reasoning models that have been specifically trained on primary care clinical data, rather than the general-purpose models that were assessed in this study.

---

Table 1: Results of each model's performance

| Question | o3 | Claude Opus 4 | Gemini 2.5 Pro | Grok-3 |
|---|---|---|---|---|
| 1 | Correct | Correct | Correct | Correct |
| 2 | Correct | Correct | Correct | Correct |
| 3 | Correct | Correct | Correct | Correct |
| 4 | Correct | Correct | Correct | Correct |
| 5 | Correct | Incorrect | Correct | Correct |
| 6 | Correct | Correct | Correct | Correct |
| 7 | Correct | Correct | Correct | Correct |
| 8 | Correct | Correct | Correct | Correct |
| 9 | Correct | Correct | Correct | Correct |
| 10 | Correct | Correct | Correct | Correct |
| 11 | Correct | Correct | Correct | Correct |
| 12 | Correct | Correct | Correct | Correct |
| 13 | Correct | Correct | Correct | Correct |
| 14 | Correct | Correct | Correct | Correct |
| 15 | Correct | Incorrect | Correct | Correct |
| 16 | Correct | Correct | Correct | Correct |
| 17 | Correct | Correct | Correct | Correct |
| 18 | Correct | Correct | Correct | Correct |
| 19 | Correct | Correct | Correct | Correct |
| 20 | Correct | Correct | Correct | Correct |
| 21 | Correct | Correct | Incorrect | Correct |
| 22 | Correct | Correct | Correct | Correct |
| 23 | Correct | Correct | Correct | Correct |
| 24 | Correct | Correct | Correct | Correct |
| 25 | Correct | Correct | Correct | Correct |
| 26 | Correct | Correct | Correct | Correct |
| 27 | Correct | Incorrect | Incorrect | Incorrect |
| 28 | Correct | Correct | Correct | Correct |
| 29 | Correct | Correct | Correct | Correct |
| 30 | Correct | Correct | Correct | Correct |
| 31 | Correct | Correct | Correct | Correct |
| 32 | Correct | Correct | Correct | Correct |
| 33 | Correct | Correct | Correct | Correct |
| 34 | Correct | Correct | Correct | Correct |
| 35 | Correct | Correct | Correct | Correct |
| 36 | Correct | Correct | Correct | Correct |
| 37 | Correct | Correct | Correct | Correct |
| 38 | Correct | Correct | Correct | Correct |
| 39 | Correct | Correct | Correct | Correct |
| 40 | Correct | Correct | Correct | Correct |
| 41 | Correct | Correct | Correct | Correct |
| 42 | Correct | Correct | Correct | Correct |
| 43 | Correct | Correct | Correct | Correct |
| 44 | Correct | Correct | Correct | Correct |
| 45 | Correct | Correct | Correct | Correct |
| 46 | Correct | Correct | Correct | Correct |
| 47 | Correct | Correct | Correct | Correct |
| 48 | Correct | Correct | Correct | Correct |
| 49 | Correct | Correct | Correct | Correct |
| 50 | Correct | Correct | Correct | Correct |
| 51 | Correct | Correct | Correct | Correct |
| 52 | Correct | Correct | Correct | Correct |
| 53 | Correct | Correct | Correct | Correct |
| 54 | Correct | Correct | Correct | Correct |
| 55 | Correct | Correct | Correct | Correct |
| 56 | Correct | Incorrect | Correct | Correct |
| 57 | Correct | Correct | Correct | Correct |
| 58 | Correct | Correct | Correct | Correct |
| 59 | Correct | Correct | Correct | Correct |
| 60 | Correct | Correct | Correct | Correct |
| 61 | Incorrect | Correct | Incorrect | Correct |
| 62 | Correct | Correct | Correct | Correct |
| 63 | Correct | Correct | Correct | Correct |
| 64 | Correct | Correct | Correct | Correct |
| 65 | Correct | Correct | Correct | Correct |
| 66 | Correct | Correct | Correct | Correct |
| 67 | Correct | Correct | Correct | Correct |
| 68 | Correct | Correct | Correct | Correct |
| 69 | Correct | Correct | Incorrect | Correct |
| 70 | Correct | Correct | Correct | Correct |
| 71 | Correct | Correct | Correct | Incorrect |
| 72 | Correct | Correct | Correct | Correct |
| 73 | Correct | Correct | Correct | Correct |
| 74 | Correct | Correct | Correct | Correct |
| 75 | Correct | Correct | Correct | Correct |
| 76 | Correct | Correct | Correct | Correct |
| 77 | Correct | Correct | Correct | Correct |
| 78 | Correct | Correct | Correct | Correct |
| 79 | Correct | Correct | Correct | Correct |
| 80 | Correct | Correct | Correct | Correct |
| 81 | Correct | Correct | Correct | Correct |
| 82 | Correct | Correct | Correct | Correct |
| 83 | Correct | Correct | Correct | Correct |
| 84 | Correct | Correct | Correct | Correct |
| 85 | Correct | Correct | Correct | Correct |
| 86 | Correct | Correct | Correct | Incorrect |
| 87 | Correct | Correct | Correct | Correct |
| 88 | Correct | Correct | Correct | Correct |
| 89 | Correct | Correct | Correct | Correct |
| 90 | Correct | Correct | Correct | Correct |
| 91 | Correct | Correct | Correct | Correct |
| 92 | Correct | Correct | Correct | Correct |
| 93 | Correct | Correct | Correct | Correct |

| 94 | Correct | Correct | Correct | Incorrect |
| --- | --- | --- | --- | --- |
| 95 | Correct | Correct | Correct | Correct |
| 96 | Correct | Correct | Correct | Correct |
| 97 | Correct | Correct | Correct | Correct |
| 98 | Correct | Correct | Correct | Correct |
| 99 | Correct | Incorrect | Incorrect | Incorrect |
| 100 | Correct | Correct | Correct | Correct |
| **Total** | **99/100** | **95/100** | **95/100** | **95/100** |
| | **99.0%** | **95.0%** | **95.0%** | **95.0%** |